\documentclass[sigconf]{acmart}
\usepackage{balance}
\usepackage[most]{tcolorbox}
\usepackage{xcolor}

\setcopyright{none}
\renewcommand\footnotetextcopyrightpermission[1]{}

\AtBeginDocument{%
  \providecommand\BibTeX{{%
    \normalfont B\kern-0.5em{\scshape i\kern-0.25em b}\kern-0.8em\TeX}}}

\newif\ifano
\anofalse % toggle with \anofalse / \anotrue
\newcommand{\pill}[3]{%
  \tcbox[
    on line,
    colback=#1,      % background color
    colframe=#1,     % border color
    boxrule=0pt,
    arc=3pt,         % rounded corners
    boxsep=0.5pt,
    left=4pt, right=4pt, top=2pt, bottom=2pt,
    nobeforeafter,
    fontupper=\sffamily\bfseries\color{#2}
  ]{#3}%
}

\newcommand{\circmark}[1]{%
\tcbox[
    on line,
    colback=black,
    colframe=black,
    boxrule=0pt,
    arc=4pt,           % small arc for slight rounding
    boxsep=0pt,
    left=2pt, right=2pt, top=2pt, bottom=2pt,
    nobeforeafter,
    fontupper=\sffamily\bfseries\color{white}
  ]{#1}%
}

\definecolor{colorregular}{RGB}{137,137,137}

\definecolor{colorhuman}{RGB}{255,191,142}

\definecolor{colordecision}{RGB}{70,70,70}

\definecolor{colorllm}{RGB}{142,152,255}

\begin{document}

\title{Empowering Computing Education Researchers Through LLM-Assisted Content Analysis}

\author{Laurie Gale}
\affiliation{%
  \department{Raspberry~Pi~Computing~Education Research Centre}
  \institution{University of Cambridge}
  \city{Cambridge}
  \country{UK}
}
\email{lpg28@cst.cam.ac.uk}

\author{Sebastian M. Nicolajsen}
\email{sebni@itu.dk}
\affiliation{%
    \department{Center for Computing Education Research}
    \institution{IT University of Copenhagen}
    \city{Copenhagen}
    \country{Denmark}
}

\begin{abstract}
Computing education research (CER) is often instigated by practitioners wanting to improve both their own and the wider discipline's teaching practice. However, the latter is often difficult as many researchers lack the colleagues, resources, or capacity to conduct research that is generalisable or rigorous enough to advance the discipline. As a result, research methods that enable sense-making with larger volumes of qualitative data, while not increasing the burden on the researcher, have significant potential within CER.  

In this discussion paper, we propose such a method for conducting rigorous analysis on large volumes of textual data, namely a variation of LLM-assisted content analysis (LACA). This method combines content analysis with  the use of large language models, empowering researchers to conduct larger-scale research which they would otherwise not be able to perform. Using a computing education dataset, we illustrate how LACA could be applied in a reproducible and rigorous manner. We believe this method has potential in CER, enabling more generalisable findings from a wider range of research. This, together with the development of similar methods, can help to advance both the practice and research quality of the CER discipline.
\end{abstract}

%TODO: Add CSS concepts here----------------------------
\begin{CCSXML}
<ccs2012>
   <concept>
       <concept_id>10003456.10003457.10003527</concept_id>
       <concept_desc>Social and professional topics~Computing education</concept_desc>
       <concept_significance>500</concept_significance>
       </concept>
   <concept>
       <concept_id>10002944.10011123.10010912</concept_id>
       <concept_desc>General and reference~Empirical studies</concept_desc>
       <concept_significance>300</concept_significance>
       </concept>
   <concept>
       <concept_id>10002944.10011123.10010916</concept_id>
       <concept_desc>General and reference~Measurement</concept_desc>
       <concept_significance>300</concept_significance>
       </concept>
   <concept>
       <concept_id>10010147.10010178</concept_id>
       <concept_desc>Computing methodologies~Artificial intelligence</concept_desc>
       <concept_significance>500</concept_significance>
       </concept>
 </ccs2012>
\end{CCSXML}

\ccsdesc[500]{Social and professional topics~Computing education}
\ccsdesc[300]{General and reference~Empirical studies}
\ccsdesc[300]{General and reference~Measurement}
\ccsdesc[500]{Computing methodologies~Artificial intelligence}

\keywords{Content analysis, large language models, qualitative data, computing education research methodology}
%-----------------------------------------------

\maketitle

\pagestyle{plain}%For arXiv submission

\section{Introduction}

\begin{quote}
    \textit{She closed her laptop and took a deep breath. She was certain the new approach to teaching CS1 had improved the course, but how could she convince the university? It would mean asking for more resources, and that required more than just conviction. Hard numbers would have helped, but the grade data was still weeks away. All she had now was the tangle of student comments from the course evaluations—subjective, scattered, and stubbornly resistant to quick conclusions, something which she did not have time for.}
\end{quote}

% She closed down her laptop and took a deep breath. She is about to report the effects of her latest resourceful intervention in her CS1 course to the study board. The grades are still not in. She thought to herself, could I not have made the time to conduct a more rigorous analysis of her student feedback? Is she even certain that the intervention resulted in better learning? As she enters the meeting room, and prepares her laptop, she reflects, will I ever be able to publish this?

\noindent Computing education (CE) and its associated research (CER) is a fast-growing independent discipline. Over the last few decades, we have strived to achieve better teaching, enabling more individuals to succeed in and outside of computing through computational means, as evident from the significant role which teaching, learning, course design, and student experiences play in our research \cite{simon2020twenty, becker201950}. However, CER remains a young field which, in many ways, is still maturing. A large corpus of CER studies remain to be experience reports, also spoken of as  \textit{Genesis} or \textit{Marco Polo} papers --- \textit{``And he saw what he had made, and it was good''}, respectively,  \textit{``I went there and I saw this''} \cite{simon2020twenty}. 

As our fictitious introductory story tells, proper analysis of teaching interventions, and other data collected in CER, is often a luxury. Small teams, a lack of research connections, or a primary allocation to teaching activities often make rigorous analysis unfeasible. However, through proper, rigorous analyses of the many data points we often aggregate in our research --- such as student feedback, video recordings, or log data --- we can improve not only our understanding of their effects in the classroom, but also our report to the research community. This is more important now than ever; with an increasing number of students being exposed to computing, there is a need for repeatable and generalisable research which allows us to scale and improve computing education across education levels and institutions.

With the increasing use of large language models (LLMs) in qualitative research across domains, we stand with a new opportunity as computing education teachers and researchers. More in-depth analyses can now be conducted with fewer resources, so long as LLMs are used carefully and intentionally. This discussion paper proposes LLM-assisted content analysis (LACA), which harnesses the power of LLMs within the broader method of content analysis. Specifically, LLMs are used for performing deductive coding, with the researcher responsible for deductively or inductively generating the codebook. To show a potential application of LACA, we provide an example case study with a computing education dataset. The use of LACA empowers researchers to analyse large sets of textual data that they would not otherwise be feasible. This particularly benefits computing educators lacking the ability to rigorously investigate the effect of their intervention and researchers analysing large bodies of student or teacher text with content analysis.

We believe this to be particularly important now, as existing overarching methods for LLM use in qualitative methods are lacking, making it easy and tempting for researchers to not disclose their usage. Thus, if we want to conduct trustworthy, repeatable, and ethical research with LLMs, adherence to rigorous research methods is vital. LACA is a step towards this.

\section{Content Analysis}
Content analysis (CA) is a method for analysing textual data that has developed over many decades. It has since expanded into a large umbrella of methods which have been frequently used to analyse a range of textual and non-textual data. Within CER, CA has been applied to a wide range of rich data types, addressing a vast portfolio of different research questions ( e.g., \citep{ContentAnalysisProgrammingErrors, InvestigatingPreexistingDebuggingTraits, CSPCKConceptualisation}). Before explaining \textit{how} LLMs can be used in CA, we first define what CA is and justify \textit{why} LLMs are appropriate to use. 

\subsection{Definitions and Key Principles}
We use \citet{ContentAnalysisIntroMethodology} and \citet{ContentAnalysisGuidebook} as well-respected CA guides for situating and developing LACA. They define content analysis as the following\footnote{Emphasis by the authors of this paper.}: \\

\noindent \textit{Content analysis is a research technique for making \textbf{replicable} and \textbf{valid} inferences from \textbf{texts (or other meaningful matter)} to the contexts of their use.} - Krippendorff  \citep[p. 18]{ContentAnalysisIntroMethodology} \\

\noindent \textit{Content analysis is a \textbf{summarizing}, \textbf{quantitative} analysis of messages that \textbf{follows the standards of the scientific method} (including attention to objectivity--intersubjectivity, a priori design, reliability, validity, generalizability, replicability, and hypothesis testing based on theory) and is not limited as to the types of variables that may be measured or the context in which the messages are created or presented.} - Neuendorf  \citep[p. 22]{ContentAnalysisGuidebook} \\

\noindent Based on these, there are some key principles of CA, which are described in more detail in \citeauthor{ContentAnalysisGuidebook}'s six-part definition of CA.

First, content analysis is concerned with \textbf{analysing messages}. These messages do not have to be textual but must contain some ``meaning'' \citep{ContentAnalysisGuidebook, ContentAnalysisIntroMethodology}, allowing a wide range of media to be analysed. Within CER, this has enabled CA to be performed on a wealth of textual and non-textual data, such as interview transcripts \citep{CSPCKConceptualisation, IntersectionalExperiencesBlackWomenComputing}, video recordings of students \citep{InvestigatingPreexistingDebuggingTraits}, and programming data \citep{ContentAnalysisProgrammingErrors, NovicesFirstLineCode, FunctionsMisconceptionsAndSelfEfficacy}.

Second, content analysis was originally developed as a \textbf{quantitative} method for \textbf{summarising} data. The typical output of content analyses is a codebook containing counts for different categories. These can facilitate further statistical analysis, comparison with related work, or greater understanding of a phenomenon. Qualitative versions of content analysis methods are now widely accepted and used (.e.g, \citep{QualContentAnalysis}), which differ in how the codebook is produced. This is common in CER (e.g., \citep{ContentAnalysisProgrammingErrors, InvestigatingPreexistingDebuggingTraits, CSPCKConceptualisation}), perhaps due to the lack of preexisting CER to base codebooks on.

For codebooks derived using CA to be considered useful, content analyses must be conducted with sufficient rigour and reproducibility. This brings us to arguably the most important principle of CA: it \textbf{abides by the principles of the scientific method}. This involves using hypothesis testing and prior theory where appropriate, so as to build on previous work, and ensuring suitable validity and reliability, such as reaching acceptable interrater reliability values. All of this should be sufficiently detailed to allow others to replicate, which is often lacking in CER and beyond.

\subsection{The Limitations of Content Analysis}
Based on these principles, a key goal of CA is to generalise findings to a wider population than the sample in question \citep{ContentAnalysisGuidebook}. This requires appropriate sampling as well as sufficient validity and reliability. Over time, repeatedly conducting larger-scale and more generalisable content analyses brings about quicker and more significant advances in the research field, improving our understanding and benefitting practice.

Unfortunately, conducting generalisable CA is often difficult in CER. First, computing education researchers often lack the personnel to conduct sufficiently large content analyses --- here the number of coders, rather than the amount of data, is typically a factor limiting the volume of data that can be coded. Second, researchers may lack the time to conduct sufficiently rigorous CA, especially if they are analysing other data in the same study. Even if there were sufficient time and resources, coordinating large-scale content analyses adds an organisational burden and becomes harder to ensure reliability.

\subsection{\textit{Why} LLMs Can Help}
We believe the use of LLMs in CA can enable researchers and educators to perform analysis that they would not otherwise be able to perform. Just as importantly, we believe that LLMs can be used in a methodologically sound manner that conforms to the principles of CA for several reasons.

First, the principles of the scientific method, particularly reliability, validity, and replicability, can be adhered to when using LLMs in CA. During the process of deductive coding, LLMs can effectively replace the role of human coders, making measures of reliability and validity verifiable in the same way. Interrater reliability (IRR) measures, for example, function the same regardless of how the codes for a set of data were generated. To facilitate replicability, details of the LLM can be reported, such as the model, prompt, and dates of experiments \citep{LLMQualAnalysisUsesTensions, LLMEmpiricalStudyGuidelines}.

Additionally, the task of deductive coding in content analysis suits the capabilities of LLMs. Codebooks should contain instructions, examples, definitions, and anything else that allows human coders to reliably code the same data \citep{ContentAnalysisGuidebook}. These elements of a clearly defined codebook also enable LLMs to reliably code data, especially when combined with appropriate prompt engineering techniques \citep{DeductiveCodingAIHumanPerformance, PotentialAndLimitsLLMsQualCoding}. There is already some evidence to suggest that LLMs can perform such deductive coding with good interrater reliability \citep{LACAPaper, DeductiveCodingAIHumanPerformance, QualCodingGPT4} and do not perform well on categories that humans struggle to reliably code \citep{QualCodingGPT4, PotentialAndLimitsLLMsQualCoding}. A consequence of this is arguably positive; LLMs will not code as reliably if the codebook provided is not sufficiently detailed, which encourages researchers to use clearly defined codebooks.

Not only can LLMs follow clear human instructions, the summarising nature of CA is well suited to LLMs. While some interpretation may be required, CA is `less in-depth and detailed' in the interest of reliability and generalisability \citep{ContentAnalysisGuidebook}. Performing coding where LLMs are required to perform interpretation or have more contextual knowledge can yield unreliable results \citep{PotentialAndLimitsLLMsQualCoding}, which can also be the case with humans \citep{ContentAnalysisGuidebook}. This is not useful when trying to assess counts or perform other statistical analyses.

\subsection{\textit{How} LLMs Can Help}\label{sec:how-llm-help}
Based on the considerations in the previous sections, we believe LLMs best serve as deductive coding agents in CA. This application of LLMs serves several benefits, particularly for researchers lacking time, colleagues, or resources.

One obvious advantage of using LLMs is scale. In traditional CA, the number of researchers limits the sample of texts that can be analysed \citep{ContentAnalysisIntroMethodology}. As long as the principles of CA are followed, it can be applied to very large datasets; one example in \citet{ContentAnalysisGuidebook} includes a CA of 31 million words involving 31 coauthors/coders. Provided with a suitable codebook, LLMs can perform vast amounts of coding, hugely expanding the amount a single or small group of researchers can analyse. No longer is the number of researchers the limit for the amount of data CA can be performed on.

An associated benefit is speed. Even if one did have a group of 31 trained coders, the analysis would take time to perform. This is not ideal for quick scoping analyses or researchers pressured by time, perhaps due to teaching commitments. LLMs, on the other hand, can perform deductive coding much quicker than humans. Where justification is required to resolve disagreements and improve reliability, chain-of-thought reasoning can also help to identify coding patterns in the LLM \citep{LLMsCanIntroduceSeriousBias}.

% The example does not try to contribute in new ways, it is trying to illustrate the usefulness of the method. 

% The particular reason this example is chosen is because the data set has grown to the size it has, and humans can no longer do this manually.

% We exemplify use-case (add to introduction and abstract)

% - There's been a line of work for the last 10-15 years that's explored CER's growth as a discipline conducted by Simon/Malmi/Sheard etc. This includes domain-specific theories, use of methods, and other analysis of CER papers. As a result, this dataset relates to the research area around the growth and development of the CER discipline, which is useful for tracking how our field is evolving.
% - What's more, lots of this work USES content analysis to conduct their analysis!
% -It's available: While course evaluations/something similar would've been nice, we struggled to find a relevant open dataset, which I think is reasonable justification to go with this one

% In other words, the example relates to an existing line of work which has utilised CA, which LLMs can help to expand on (due to the increasing number of papers).

% Then perhaps we can say that other useful datasets include short-medium textual data e.g. student course evaluations and teacher perspectives (with appropriate references)

\section{LLM-Assisted Content Analysis}\label{sec:process-and-example}
We now propose \textit{LLM-assisted content analysis (LACA)}, an instance of content analysis that incorporates LLMs for the purposes of deductive coding. This builds on an existing method of the same name \citep{LACAPaper}, with extensions related to methodological and ethical rigour. Unlike other aspects of qualitative and content analyses, we believe the use of LLMs is methodologically compatible with and technically proficient at the deductive aspect of content analysis. Conversely, we do not believe LLMs can effectively perform more interpretive forms of qualitative analysis or inductive code generation.

%Before we introduce the method, we must make our beliefs around what LLMs can reliably be used to perform clear. First, we do not believe LLMs can effectively perform more interpretive forms of qualitative analysis or inductive code generation. %As well as previous work suggesting poor results with inductive qualitative analysis (refs), such analysis requires a good understanding of the research questions and immersion in the data. These research questions will evolve based on the analysis, requiring coders to be adaptable and reflexive. We believe such creativity in analysis is unique to humans.

\begin{figure}
    \centering
    \includegraphics[width=\linewidth]{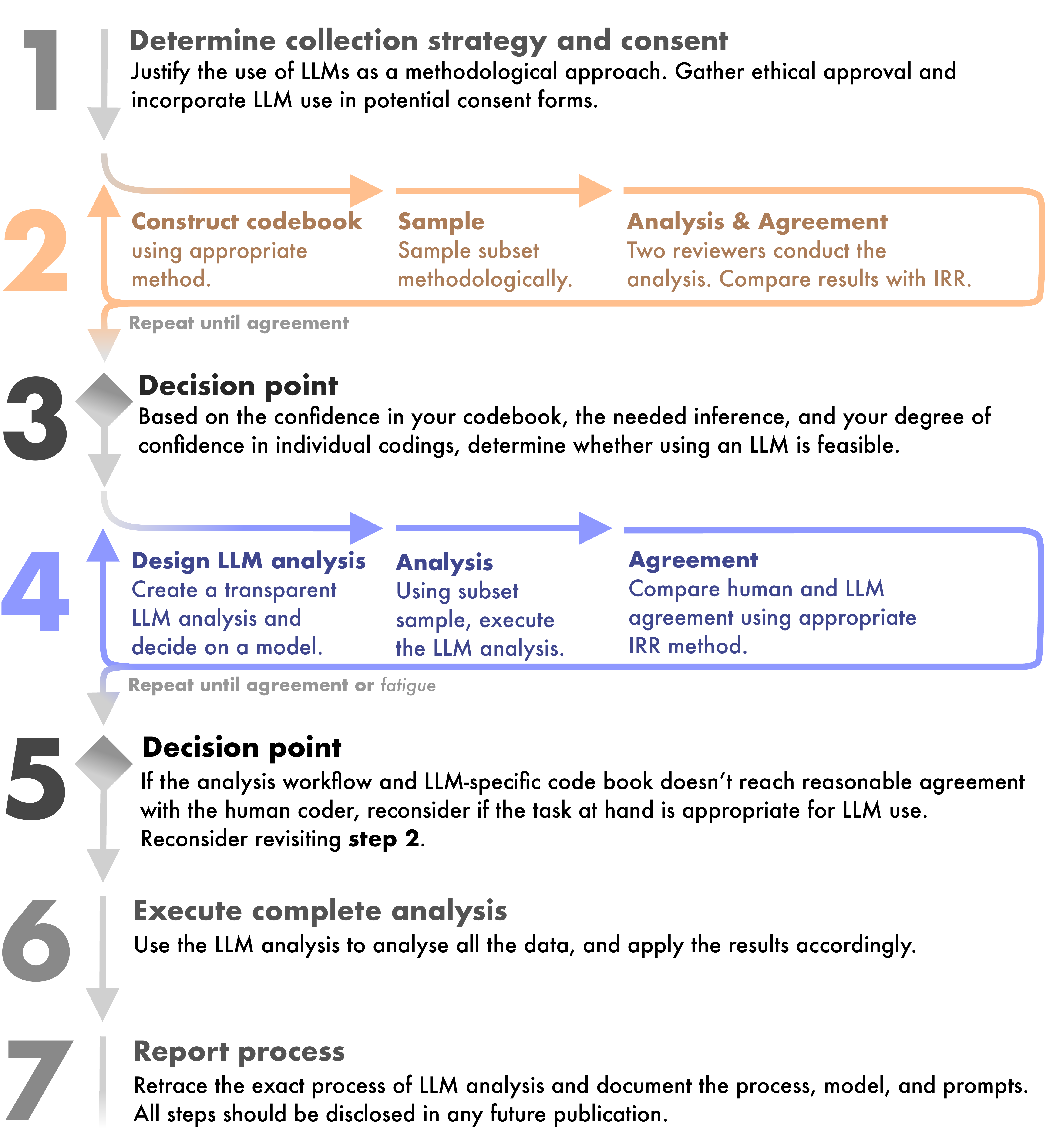}
    \caption{The LACA Process}
    \label{fig:laca-process}
\end{figure}

Figure \ref{fig:laca-process} visualises the LACA process. 
In this section, we outline the process and exemplify the individual steps using a fictitious worked example. While the example is not complete, we provide all the necessary documentation for conducting it in an online repository \cite{lacarepo}. 

With LACA, we highly emphasise how the steps prior to the use of an LLM are crucial in ensuring proper, ethical application. In particular, researchers should consider whether applying LACA is a suitable approach to use, and inform potential participants and ethical committees. Generally, we recommend local models, in accordance with ethical recommendations \cite{LLMQualAnalysisUsesTensions}, especially if the application of LACA is decided post data collection.

The example we demonstrate LACA on is a continuation of \citeauthor{simon2020twenty}'s research, who sought to analyse two decades of papers at the \textit{Innovation and Technology in Computer Science Education (ITiCSE) conference}. The data set includes abstracts from publications from SIGCSE TS (7936), ITiCSE (3154), ICER (730), TOCE (334), Koli Calling (316), GCE (103). The data set was aggregated using Semantic scholar.\footnote{\url{https://www.semanticscholar.org}} The full data set is accessible through the link to our example \cite{lacarepo}.

This research relates to a larger body of work that has explored CER's growth as a discipline, including the use theory, methods and statistics (e.g., \citep{simon2007classification, AttitudesEmotionsLitReview, DemographicsInCER, InferentialStatsCER}). Many of these studies have used CA to conduct their analysis, which will naturally become harder as the CER discipline grows. We therefore use this example due to the number of publications outweighing the capability of human coding involved and the dataset being easily acquirable. From here on, text related to this example will appear in italics. \\

%\vspace{1em}
%\noindent\textit{The example we will work through during the exploration of the LACA method, is a (limited) continuation of the work presented by Simon \& Sheard \cite{simon2020twenty}. In particular, we will analyse \textbf{themes} in computing education publications in a data set of 12,573 computing education abstracts aggregated through semantic scholar.\footnote{The dataset includes publications from SIGCSE TS (7936), ITiCSE (3154), ICER (730), TOCE (334), Koli Calling (316), GCE (103). The full data set is accessible through the link to our example \cite{lacarepo}. We chose the collection of CER paper abstracts as our example due to its size, which exceeds what can be analysed manually, making automated methods necessary. We also chose this particular example as it is tied to a long line of research in the community, which includes, but is not limited to, the work by Simon \& Sheard \cite{simon2020twenty, simon2007classification}.} All example parts will appear in italic like here.} 

\noindent \pill{colorregular}{white}{Step 1} First, one should consider whether LACA is an appropriate choice of method for the research to be conducted. This includes considering the complexity of the data aggregated and whether alternative types of analysis may perform better. One should also consider the model of choice; local models increase privacy while remote models may collect data and be more expensive. However, large local LLMs may take significant time to complete generating responses, depending on the machine available.\\

\noindent\textit{In the case of our example, we are analysing abstracts, looking for certain themes emerging. Here, we first consider whether a simple analysis based on word occurrence will suffice. However, after manual exploration of some of the data, we found that different abstracts utilise a wide array of words to describe different concepts, which we cannot guarantee to catch by simply utilising word lists. We also realise that the complexity of abstracts can be significant, and thus opt for one of the larger local LLMs. In this case, we use \textbf{gemma 3 27B} as this can be run on a mid-range MacBook Pro M3. We do not consider any anonymisation procedures as the data under investigation will contain no sensitive data.} 
\\

\noindent \pill{colorhuman}{white}{Step 2} Then, one should construct the codebook. It is important to state that the LACA process does not provide any particular means to do this. Instead, researchers should opt for a method of codebook generation which is appropriate for the research. Second, one should consider a representative sample to execute the preliminary \textit{human} review of and decide on an appropriate IRR measure to use, such as Cohen's $\kappa$ \citep{IRRKappaStatistic}, Krippendorff's $\alpha$ \citep{KrippendorffComputingAlpha-Reliability}. Two or more reviewers should then conduct the content analysis based on the designed codebook, and iteratively improve the codebook to achieve an acceptable level of inter-rater reliability. \\

\noindent\textit{Given that we are extending the work by Simon \& Sheard \cite{simon2020twenty}, we opt to design the codebook for this example by reusing their description of themes, and including the themes which they identify in the study, as most prominent \cite{lacarepo}. For the sake of illustration, the themes include, but are not limited to; ``Teaching/learning techniques'', ``Teaching/learning tools'',  ``Recruitment, progression, pathways'', and ``Gender issues''. Using this, two reviewers would code a random sample of 1,257 abstracts (10\%) to determine the quality of the codebook. Discussing disagreements, we would revise the codebook and continue until inter-rater reliability was high enough to constitute agreement. For this, we decide on utilising a variation of Krippendorff's $\alpha$ which allows each reviewer to assign multiple codes to a single data point, and strive for an $\alpha$-value above 0.80, as recommended by \citet{ContentAnalysisIntroMethodology}} 
\\

\noindent \pill{colordecision}{white}{Step 3} Based on your findings from \pill{colorhuman}{white}{step 2}, you should consider whether LACA is still appropriate. It can, therefore, be appropriate to include additional information during the human coding. This could include comparative indicators such as how weak or strong individual codings are compared to others, to help assert the appropriateness of the method, as described in Section \ref{sec:how-llm-help}.

\vspace{1em}
\noindent\textit{Given the fictitious nature of our example, we assume a Krippendorff's $\alpha$ higher than our threshold of 0.80 was achieved through dialogue about disagreements and iterations on the codebook. Given the directness of the themes and the high IRR, we assume continuing with a large context local LLM is feasible.} 
\\

\noindent \pill{colorllm}{white}{Step 4} The next step is employing an LLM to conduct the analysis on the same sample as the human coders, and then, calculate the IRR between human and LLM. The prompt from which we start is simply the codebook used in the previous stages. If the IRR achieved initially is not above the selected threshold, one should refine the codebook (now prompt) by inspecting the resulting analysis. It is here important to recognise that, despite laudable efforts and many iterative cycles, the LLM may never reach IRR which is above the pre-defined acceptable threshold. We refer to this as \textit{\textbf{fatigue}}. Thus, it is important to monitor whether the IRR stagnates despite changes to the prompt. In these cases, one should reconsider \textit{not} moving forward with LACA.

If one does continue, there are many ways to design an analysis `flow' that illustrates the LACA process in a transparent and repeatable manner. For this purpose, we developed a small \textit{library} \cite{aitomics} and no-code solution \cite{aitomicsnocode} which allows users to conduct LACA programmatically. In particular, the library is designed to record the flow of analysis, so as to increase transparency of what transformations the data goes through.

\begin{figure*}
    \centering
    \includegraphics[width=1\linewidth]{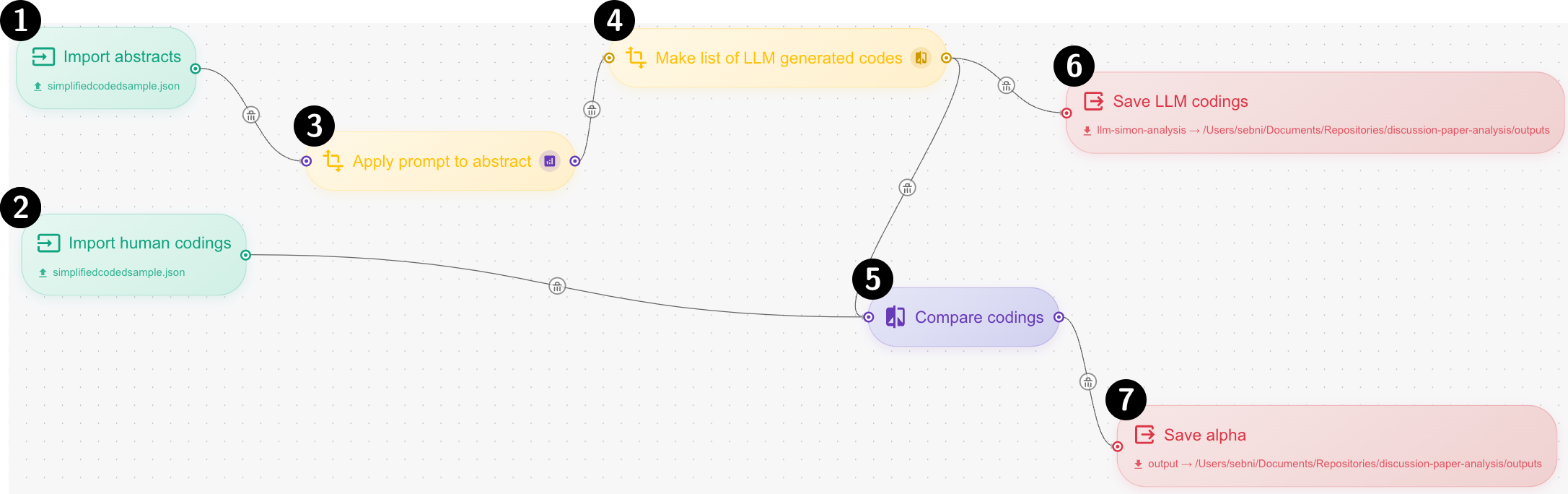}
    \caption{Illustration of LLM workflow in our custom no-code tool for doing programmatic analysis with local LLMs.}
    \label{fig:example-workflow}
\end{figure*}

\vspace{1em}
\noindent\textit{To conduct the LLM analysis on the sample coded by humans, we design a workflow in our visual tool \cite{aitomicsnocode}, which automatically produces a shareable codebook and allows us to compare the human-coded data, saved as a JSON file, with that generated by the LLM, saved as a CSV file.  We make some minor changes to the codebook (see \cite{lacarepo}), so that we can more easily execute the modified Krippendorff's alpha previously mentioned (also implemented by the tool). Figure \ref{fig:example-workflow} illustrates the specific flow. First, the abstracts \circmark{1} and human codes \circmark{2} are imported. Then, we apply the LLM prompt on each abstract \circmark{3} and modify the output to make it comparable to the human codes \circmark{4}. We then compare the human codes to the LLM codes using the chosen IRR measure \circmark{5}, and save both the LLM generated codes \circmark{6} and the IRR value \circmark{7}. We then simply run the flow, modify the prompt as needed depending on the IRR produced, and continue until acceptable IRR is achieved on the sample.} 

\vspace{1em}
\noindent \pill{colordecision}{white}{Step 5} Based on whether the LLM is capable of reaching acceptable IRR with the human codings, decide on the next action. If the IRR is acceptable, you move on to \pill{colorregular}{white}{step 6}. If not, reevaluate whether LACA is appropriate, potentially returning to \pill{colorhuman}{white}{step 2} to revisit your codebook (not prompt), design, and choice of LLM.

\vspace{1em}
\noindent\textit{Based on our fictitious example, we assume that after a series of iterations, the prompt achieves acceptable IRR, and we therefore continue to \pill{colorregular}{white}{step 6}.} 
\\

\vspace{1em}
\noindent \pill{colorregular}{white}{Step 6} Now, utilise the LLM to code the entire dataset. The analysis flow should be equivalent to the analysis in \pill{colordecision}{white}{Step 5} and be reflected in any reporting of LACA. %The generated analysis of the full data set can then be applied as the foundation of your analysis. 

\vspace{1em}
\noindent\textit{In the case of our example, the analysis flow, generated from our tool, is run on the entire data set,  excluding the comparison component (see \cite{lacarepo}).}
\\

\vspace{1em}
\noindent \pill{colorregular}{white}{Step 7}  In line with existing guidelines on reporting LLM use in qualitative research \citep{LLMQualAnalysisUsesTensions} and research generally \citep{LLMEmpiricalStudyGuidelines}, it is crucial to report details regarding the LLM used and the particular choices made within the LACA process. There are many things one could report. We recommend that one reports; 1) that an LLM was used to perform content analysis; 2) the choice of model and version (i.e., open-source or proprietary model, local or remote); 3) data anonymisation procedures; 4) Sampling sizes; 5) IRR measure used and final values for both human-human and human-LLM comparisons; and 6) the codebook and analysis flow used.

Some of this information need not documenting in the main body of the paper. More intricate details for precise replication of the study can be included as an appendix or in an open study repository.

\vspace{1em}
\noindent\textit{Based on the analysis conducted as part of our example, one could report all of the above in a similar way to the following (but including IRR values explictly):} \\
\vspace{1em}
\noindent \textit{We applied LLM-Assisted Content Analysis. First, designing a codebook using sections of \citet{simon2020twenty} (see Appendix A), and refining this based on the IRR (A modified version of Krippendorff’s $\alpha$ to support multiple codes from each reviewer) achieved by two reviewers coding a random sample of 1,257 publications (10\%). We then utilised a local Gemma 27B model to analyse each abstract in the sample, comparing it to the human codes using the same IRR method as before, and achieved a high $\alpha$-value, implying agreement between humans and the LLM ($\alpha$ > 0.80). Then, we conducted the LLM-based analysis on the entire dataset (12,573 publication abstracts), which lays the foundation of our findings. The flow of analysis is visualised in Figure 1. For the codebook and LLM analysis flow, see Appendix B.}

% Defence of the method and the tooling
% REUSING EXAMPLE IN INTRODUCTION to discuss all these things.

% Argument 1: But having the computer doing the thinking eliminates the researcher?

% Argument 2: ...

% Argument 3: ...

% Argument 4: ...

\section{Existing qualitative approaches with LLMs}

% Generally introduction to all the GREAT work going on, but also why, it isnt exactly ethical...

% Expanding on, in particular, all the negative....
% Inductive approaches
% Catch-all-solutions (black box approaches)

% We want it to be not-a-thinker an assistant to the researcher doing the high level thinking

% What we like more
% - Ethical considerations
% - LACA method

% General introduction on how this is mostly in different disciplines

% NLP vs LLMs
% ad-hoc llm use vs training classifiers

The applications of LLMs to qualitative research has unsurprisingly been widely investigated in the last few years. However, much of the work resides outside of computing education research. The research which does exist in computing education employs complex analysis such as combining multiple existing methods, including using Retrieval-Augmented Generation (RAG) to extract rationales in publications \cite{schulte2025we}. While \citet{schulte2025we} also contribute a custom library to conduct similar analysis to theirs, they do not explicitly incorporate IRR in their method, explicitly consider the ethical considerations outlined by \citet{LLMQualAnalysisUsesTensions}, or provide instructional guidelines to recreate their method.\footnote{While instructional guidelines are not expected, it underlines the need for method-oriented publications.}

Thus, we here provide a brief overview of some general areas of application, as well as some ethical and methodological objections. We do not provide a comprehensive report given the large traction LLM use has gained in qualitative research. Further, we do not report on `traditional' NLP approaches, e.g., fine-tuning of BERT models, as the purpose of this research is to make NLP approaches more accessible for researchers.

Several works have tried to integrate LLMs into existing qualitative methods to develop new methodological contributions. An example of this is LATA, or LLM-assisted thematic analysis \citep{LATA}, which aims to incorporate LLMs into the process of thematic analysis as defined by \citet{ThematicAnalysisOriginalArticle}. LLMs are involved in inductively generating open codes, searching for themes, and coding using the final set of themes. The use of LLMs application to TA has similarly been applied by \citet{SaturationThematicAnalysisLLMs}, who look to assess valid uses of LLMs. However, the process of inductive theme generation by LLMs is arguably methodologically opposed to the principles of reflexive thematic analysis. Researchers' experiences, background, and interpretations are key aspects of a TA procedure \citep{ConceptualDesignThinkingThematicAnalysis}, which LLMs do not possess. Performing interrater reliability measures, as these articles do, is also opposed to reflexive TA \citep{ConceptualDesignThinkingThematicAnalysis}. More generally, we are methodologically opposed to the use of LLMs for any inductive code generation that does not involve humans. Inductive code generation requires good understanding of the research questions and immersion in the data. These research questions will evolve based on the analysis, requiring coders to be adaptable and reflexive. We believe such creativity in analysis is unique to humans.

%Other research has more generally applied LLMs to the process of inductive code development in qualitative analysis \cite{LATA}. We are again, methodologically, opposed to this application of LLMs for reasons similar to those stated above. Inductive code generation requires good understanding of the research questions and immersion in the data. These research questions will evolve based on the analysis, requiring coders to be adaptable and reflexive. We believe such creativity in analysis is unique to humans. Another similar strand of coding is the coding of latent concepts, defined as concepts whose coding requires some interpretation. Where interpretation or context is required, reliable coding generally becomes more difficult, which is why such forms of qualitative analysis do not pose positivist constraints of reliability \citep{ThematicAnalysisOriginalArticle}. Indeed, reliably coding latent concepts has been proved difficult for humans to do \citep{ContentAnalysisGuidebook}, let alone LLMs. Although some positive results have been reported here \citep{CodingLatentConceptsLLMs}, we generally believe that these sorts of coding require interpretation that is not effectively performed by an LLM.

More work investigates the ability of LLMs to deductively code data. That is, where LLMs are provided with a codebook rather than prompted to generate one. In this case, if the codes are reliable enough, LLMs can simply replace the role of a trained coder who has not conducted any of the prior research. Initial research shows promise in terms of speed and accuracy. \citet{DeductiveCodingAIHumanPerformance} found that LLMs achieved a higher accuracy than humans when deductively coding open-text responses. Other studies have found that, for some codes, LLMs' `performance' is similar to humans \citep{LACAPaper, DeductiveCodingAIHumanPerformance, QualCodingGPT4}. However, `performance' is often measured with a confusion matrix rather than IRR values, which is more of a `traditional' NLP approach to evaluation.

% \section{Application to a Computing Education Dataset}

% Given time constraints, I wonder if it would be ok to use a fictional example, which may or may not involve mock data. The point wouldn't be to present findings, but illustrate how LACA works in practice.

% \subsection{Reproducibility}
% We followed the guidance of \citet{LLMQualAnalysisUsesTensions} and (refs)...
% \begin{itemize}
%    \item The datasets
%    \item GitHub Repo
%    \item LLM specs
%    \item LLM prompts
%\end{itemize}

% The worked examples of the dataset

% \subsection{Results}

\section{Applications, Limitations, and Reflections}
We argue that CER is a fruitful discipline for conducting LACA. Not only is content analysis frequently used within CER, many members of the CER community are intertwining their teaching with their research, or lack the resources to perform medium-large scale analysis of textual data often collected from participants. LACA provides a method for analysing such data, even when researchers do not have the capacity to manually inspect it all themselves.

The sort of data that arises from these experiences varies widely, providing many good applications of LACA within CER. For example, educators of large undergraduate courses may have a large number of students' written feedback from which they wish to identify students' struggles or conceptions (e.g., \citep{CS1NotionsCodeQuality, ComputersSpeakDifferentLanguage, LearningWebDev}). LACA could be performed to speed up the coding of this process, as well as enable aggregation of data over multiple institutions. Alternatively, CER researchers may, for example, wish to categorise students' programming data to understand common patterns in programming behaviour (e.g., \citep{ContentAnalysisProgrammingErrors, NovicesFirstLineCode, FunctionsMisconceptionsAndSelfEfficacy}). Applying LACA again increases the number of logs which could be analysed \citep{QualCodingGPT4}, complementing quantitative data obtained. As another example, tool developers may wish to analyse masses of textual student data inputted into their tool to, for example, summarise common misconceptions or demonstrate the efficacy of their tool.

The purpose of these examples is not to provide an exhaustive list of applications, but rather a starting point to motivate the range of CER that LACA enables. In doing this, however, we also acknowledge a limitation of LACA: the range of data used in CER is much more varied than the textual data that is appropriate for LACA. CER researchers have also analysed visual data \citep{AssessingUnderstandingExpressions}, video recordings of students \citep{InvestigatingPreexistingDebuggingTraits}, and even physical data from students \citep{EDAProgrammingEmotions}. Future work involves expanding the application of LACA beyond textual data to enable a wider range of large-scale analysis.

Another limitation of the current version of LACA is the lack of clarity around when to stop. While we acknowledge the possibility \textit{fatigue}, we do not know how to reliably verify whether this has happened. In other words, if a researcher(s) has already repeated \pill{colorllm}{white}{step 4} five times, should they do it again? We believe certain heuristics in future applications of LACA  will help to determine this. Either way, continuous iteration and trial-and-error situations may inadvertently make LACA more time-consuming than general content analysis (depending on the data, codebook, and utilised model).

The final limitation we note is the uncertainty of what data LACA performs well on. While our example in Section \ref{sec:process-and-example} includes relatively short-form data which is easy to code, it is not entirely clear what sort of data favours the use of LLM coding. Related work indicates that LLMs can code more concrete concepts \citep{QualCodingGPT4, DeductiveCodingAIHumanPerformance}. However, the lack of clarity around suitable data clouds the judgment required for \pill{colorregular}{white}{step 1} and \pill{colordecision}{white}{step 3}. Based on our current experiments and existing research, we suspect LLMs will perform better on short- to medium-form textual data. The longer the text, the more complex coding becomes, the more interpretation generally required, and the more margin for error. Accurate coding is likely to decrease as the length and importance of context increase.

As a consequence of our limitations, a core strand of our further work is the use of LACA in many situations where we envision it to be of benefit. As CER researchers, we wish to start with applications within our rich field. As well as enabling previously unfeasible research, each deployment of the LACA method serves as an opportunity to learn more about what sort of textual data it can be performed on. interactive rebase in progress; onto b6e3dafThis includes details about the type of data, the complexity of the codebook, and the amount of contextual knowledge required to perform coding. As we advocate the use of LLMs, results from these studies can also be compared with more traditional natural language processing techniques such as sentiment analysis. These areas of future work will inform both the technical and methodological aspects of LACA

\section{Conclusion}
Computing education research (CER) is a healthy community consisting of researchers from a wide range of disciplines, backgrounds, and experiences. However, it is still a young field, with the rigour and generalisability of findings within the field often limited by the number or capacity of people to analyse data. In this discussion paper, we have proposed a methodological solution to this problem within the context of textual data through LLM-assisted content analysis (LACA). Not only does LACA enable researchers to conduct larger-scale analysis quicker, it also encourages this to be done in a repeatable and rigorous manner. We hope this paper can be used to encourage the responsible and reliable use of LACA, and LLMs more generally, within the CER community.

\balance

\begin{acks}
Include acknowledgements if you wish
\end{acks}

%TC:ignore
\bibliographystyle{ACM-Reference-Format}
\bibliography{references-seb, references-laurie}

%TC:endignore

\end{document}